\def\year{2018}\relax
\documentclass[letterpaper]{article}
\usepackage{aaai18}  
\usepackage{times}
\usepackage{helvet}
\usepackage{courier}
\usepackage{float}
\usepackage{setspace}
\usepackage{cprotect}
\usepackage{verbatim}
\usepackage{epsfig}
\usepackage{graphicx}
\usepackage{amsmath}
\usepackage{makeidx}
\usepackage{amssymb}
\usepackage{multirow}
\usepackage{booktabs}

\title{Estimating Gradual-Emotional Behavior in One-Minute Videos with ESNs}
\author{ Tianlin Liu\\
Department of Computer Science and Electrical Engineering\\ 
Jacobs University Bremen\\
 Bremen, Germany \\
\texttt{tliu@jacobs-alumni.de}\\
\And
Arvid Kappas\\
Department of Psychology and Methods \\ 
Jacobs University Bremen\\
Bremen, Germany \\
\texttt{a.kappas@jacobs-university.de}}
\begin{document}
\maketitle

\begin{abstract}
In this paper, we describe our approach for the OMG-Emotion Challenge 2018. The goal is to produce utterance-level valence and arousal estimations for videos of approximately 1 minute length. We tackle this problem by first extracting facial expressions features of videos as time series data, and then using Recurrent Neural Networks of the Echo State Network type to model the correspondence between the time series data and valence-arousal values. Experimentally we show that the proposed approach surpasses the baseline methods provided by the organizers. 
\end{abstract}

\section{Introduction}

Modeling emotional behavior is one of the most important topics in Affective Computing (AC, see \cite{Picard1997,Calvo2015}). Numerous efforts have been dedicated to modeling emotional expressions of affective systems based on a set of basic emotions (e.g., six universal emotions in Paul Ekman's categorization scheme) or are based on instantaneous emotion categorization (e.g., features of a couple of seconds). However, researchers are still facing difficulties in modeling and analyzing long-term emotional behavior process, partially due to the lacking a suitable corpus \cite{Barros2018}. Furthermore, there are considerable doubts as to whether a discrete emotions approach with a small number of ``basic emotions'' is indeed very helpful to characterize complex affective states and hence also as diagnostic categories \cite{Kappas2010}.

In the OMG-Emotion challenge 2018, research efforts for Affective Computing models of long-term emotional behavior process are promoted. The challenge provides a dataset named One-Minute Gradual-Emotional Behavior dataset (OMG-Emotion dataset), which is a corpus with long-term emotional data together with rigorous annotations.  We decided to participate this challenge as we are interested in the computational approach to study affect, for which the challenge provides a perfect testbed.

\section{Approach}
Our architecture contains two modules. In the first module, we extract facial expression features from raw videos. In the second module, we use an Echo State Network (ESN), a variant of Recurrent Neural Networks (RNNs), to learn the coherence between facial expression and valence-arousal values. We will elaborate these two modules in the following subsections. Our initial plan was to include sentiment mining but we found that the included transcriptions contained too many errors. Other modalities were not included in our analysis. However, facial behavior has frequently been shown to be the strongest component when human raters try to describe the affective state of senders.

		\begin{table}[H]
		\small
			\begin{tabular}{*{2}{c} }
			\toprule
				Action Unit &  Description \\ 
			\midrule	
				 AU1 & Inner Brow Raiser	 \\ 
				 AU2 & Outer Brow Raiser 	 \\ 
				 AU4 & Brow Lowerer	 \\ 
				 AU5 & Upper Lid Raiser \\ 
				 AU6 & Cheek Raiser	 \\ 
				 AU7 & Lid Tightener	 \\ 
				 AU9 & Nose Wrinkler \\ 
				 AU10 & Upper Lip Raiser		 \\ 
				 AU11 & Nasolabial Deepener		 \\ 				 
				 AU12 & Lip Corner Puller	 \\ 
				 AU14 & Dimpler	 \\ 
				 AU15 & Lip Corner Depressor \\ 
				 AU17 & Chin Raiser	 \\ 
				 AU18 & Lip Puckerer		 \\ 
				 AU20 & Lip stretcher	 \\ 
				 AU23 & Lip Tightener	\\ 
				 AU24 & Lip Pressor	 \\ 
				 AU25 & Lips part	\\ 
				 AU26 & Jaw Drop	\\ 
				 AU28 & Lip Suck	\\ 
			\bottomrule
			\end{tabular}
			\caption{The 20 AUs extracted from the utterances.}
			\label{tab:AUs}
		\end{table}
		
\begin{table}[H]
	\small
			\begin{tabular}{*{2}{c} } 
			\toprule
				 ESN parameters  & Value \\ 
			\midrule
				 Number internal units & 500 \\ 
				 Spectral radius of reservoir & 1.5 \\ 
				  Ridge regression constant &  0.1 \\ 
				  Leakage & 0.85 \\ 
		     \bottomrule
			\end{tabular}
			\caption{Parameters for training the ESN.}
			\label{tab:parameters}
		\end{table}
		
\begin{table*}[t]
\small
\begin{tabular}{*{9}{c}}
\toprule
~ & \multicolumn{4}{c}{CV} & \multicolumn{4}{c}{Eval} \\
~ & \multicolumn{2}{c}{CCC} & \multicolumn{2}{c}{MSE} & \multicolumn{2}{c}{CCC} & \multicolumn{2}{c}{MSE} \\
~ & Arousal & Valence & Arousal & Valence & Arousal & Valence & Arousal & Valence  \\ 
\midrule
Vision - Face Channel \cite{Barros2016} & N/A & N/A & N/A & N/A & 0.12 & 0.23 & 0.053 & \textbf{0.12}   \\
Audio - Audio Channel \cite{Barros2016} & N/A & N/A & N/A & N/A & 0.08 & 0.10 & 0.048	 & \textbf{0.12}  \\
Audio - OpenSmile Features& N/A & N/A & N/A & N/A & 0.15	 & 0.21 & \textbf{0.045} & 0.1o \\
Text & N/A & N/A & N/A & N/A & 0.05 & 0.20 & 0.062 & 0.123 \\ 
Our Approach & 0.13 & 0.29 & 0.053 & 0.13 & \textbf{0.18} & \textbf{0.31} & 0.049 & \textbf{0.12} \\
\bottomrule
\end{tabular}
\caption{Testing results for 5-fold cross-validation (referred as CV) and for the validation set provided by the organizers (referred as Eval). \emph{CCC}: Concordance Correlation Coefficient, MSE: Mean-Squared Error. The results from the proposed method are compared with the baseline methods provided by the challenge organizers.} 
\label{tab:result}
\end{table*}

\subsection{First Module: Feature Extraction}

In this module, we extract facial expression features from videos in the utterance level. To this end, we use the Facial Action Coding System (FACS, \cite{Ekman1978}), a fully standardized classification system that codes facial expressions based on anatomic features of human faces. With the FACS, any facial expression can be decomposed into a combination of elementary components called Action Units (AUs). More specifically, we use the Emotient FACET, a computer vision program that provides frame- based estimates of the likelihood of AUs, where each frame has a duration of $1 \slash 30$s. The 20 AUs we use are listed in Table \ref{tab:AUs}. In addition to these AUs, we use neutral evidence, positive evidence, and negative evidence of each utterance as features, where the evidence is also provided by Emotient FACET. Combining AUs and the neutral, positive, and negative evidence, we have the features of 23 dimensions for each utterance. These extracted features for all utterances are available in the \texttt{sensors} folder in the package of the released codes.

\subsection{Second Module: Model Estimation}

The outputs of our first module are multi-channel time series data describing the AUs together with neutral, positive, and negative evidence for each utterance. We perform a standard pre-processing operation on the extracted features by (i) normalizing the time series data to a range between 0 and 1, and (ii) calculate the moving average of the time series to suppress the noise in the time series. Further pre-processing techniques such as PCA and ZCA whitening were also 

With the pre-processed time series data, we use Recurrent Neural Networks (RNNs) to model the time series. In particular, in this work, we use Echo State Networks (ESNs) \cite{Jaeger2004}, which is a variant of RNN, to perform the supervised learning tasks. The main idea of the supervised learning of ESNs is to (i) drive a random, large, fixed recurrent neural network with the input signal, thereby inducing in each neuron within this ``reservoir'' network a nonlinear response signal, and (ii) combine a desired output signal by a trainable linear combination of all of these response signals \cite{Jaeger2007b}. The architecture of the ESNs is ``standard'' in the sense that it adopts the classification task setup as specified in \cite{Lukovsevivcius2012} and in our previous publication \cite{Liu2018}. The hyperparameters of the ESN, tuned with 5-fold cross-validation, are listed in Table \ref{tab:parameters}.

\section{Results}		

The 5-fold cross-validation results as well as the testing results on the official validation dataset are shown in the Table \ref{tab:result}. Our result is compared with the results from four baseline methods provided by the organizers. It is encouraging that our proposed method surpasses the baseline results in the metric of CCC, which is the metric for evaluating the final challenge results. We notice that, generally, arousal values are much harder to predict than valence values for this task, which, interestingly, contrary to the findings in a related task of estimating arousal-valence values based on music \cite{Pellegrini2015}.

\section{Conclusion}
In this paper, we described our approach for the OMG-Emotion challenge 2018. Leveraging the facial expression features provided by the Facial Action Coding System (FACS), together with an Echo State Network, our model estimates the coherence between facial expression and valence-arousal values. A limitation of our method is that it only uses unimodal information, i.e., the visual information of facial expressions. For future work, it is tempting to build models which take advantage of audio, vision, and language modalities.

\bibliography{refs}
\bibliographystyle{aaai}
\end{document}